\documentclass{article}

\PassOptionsToPackage{numbers, compress}{natbib}
\usepackage[final]{neurips_2019}

\usepackage{amsmath, amsthm, amssymb, mathtools}
\usepackage{graphicx}

\usepackage[utf8]{inputenc} % allow utf-8 input
\usepackage[T1]{fontenc}    % use 8-bit T1 fonts
\usepackage{hyperref}       % hyperlinks
\usepackage{url}            % simple URL typesetting
\usepackage{booktabs}       % professional-quality tables
\usepackage{amsfonts}       % blackboard math symbols
\usepackage{nicefrac}       % compact symbols for 1/2, etc.
\usepackage{microtype}      % microtypography

\usepackage{booktabs}
\usepackage{enumitem}% http://ctan.org/pkg/enumitem

 \theoremstyle{definition}

\theoremstyle{theorem}
\newtheorem{theorem}{Theorem}[section]
 
\theoremstyle{remark}

\title{Zero-Shot Reinforcement Learning with Deep Attention Convolutional Neural Networks}

\author{%
  Sahika Genc \\
  Artificial Intelligence Lab.\\
  Seattle, WA 90121 \\
  \texttt{sahika@amazon.com} \\
  \And
 Sunil Mallya \\
  Artificial Intelligence Lab.\\
  San Francisco, CA  \\
  \texttt{smallya@amazon.com} \\
 \And
  Sravan Bodapati \\
  Artificial Intelligence Lab.\\
  Seattle, WA 90121 \\
  \texttt{sravanb@amazon.com} \\
 \And
  Tao Sun \\
  Artificial Intelligence Lab.\\
  Seattle, WA 90121 \\
  \texttt{taosun@amazon.com} \\
  \And
  Yunzhe Tao \\
  Artificial Intelligence Lab.\\
  Seattle, WA 90121 \\
  \texttt{taosun@amazon.com} \\
}

\usepackage{graphicx}

\begin{document}

\maketitle

%===============================================================================

\begin{abstract}
Simulation-to-simulation and simulation-to-real world transfer of neural network models have been a difficult problem. To close the reality gap, prior methods to simulation-to-real world transfer focused on domain adaptation, decoupling perception and dynamics and solving each problem separately, and randomization of agent parameters and environment conditions to expose the learning agent to a variety of conditions. While these methods provide acceptable performance, the computational complexity required to capture a large variation of parameters for comprehensive scenarios on a given task such as autonomous driving or robotic manipulation is high. Our key contribution is to theoretically prove and empirically demonstrate that a deep attention convolutional neural network (DACNN) with specific visual sensor configuration performs as well as training on a dataset with high domain and parameter variation at lower computational complexity. Specifically, the attention network weights are learned through policy optimization to focus on local dependencies that lead to optimal actions, and does not require tuning in real-world for generalization. Our new architecture adapts perception with respect to the control objective, resulting in zero-shot learning without pre-training a perception network. To measure the impact of our new deep network architecture on domain adaptation, we consider autonomous driving as a use case. We perform an extensive set of experiments in simulation-to-simulation and simulation-to-real scenarios to compare our approach to several baselines including the current state-of-art models. 
\end{abstract}
\providecommand{\keywords}[1]{\textbf{\textit{Index terms---}} #1}

%===============================================================================

% We describe a deep attention convolutional neural network (DACNN) architecture to address robust simulation-to-simulation (sim2sim) and simulation-to-real (sim2real) challenges for zero-shot deep reinforcement learning. 
\section{Introduction}
Most of the recent examples in deep reinforcement learning of autonomous control agents utilize realistic simulation environments to learn various tasks including but not limited to locomotion, motion planning, and robotic-arm manipulation with limited or no human guidance (see \cite{Sim-to-real:Coumans} and references therein). These realistic simulation environments are safe for the agent to experience both desired and unwanted behavior. On the other hand, in general, a controller learned in a simulation environment performs poorly in the real world or does not generalize without additional tuning in the real world.

There is no single approach for zero-shot reinforcement learning of a robotic controller agent. In \cite{DBLP:journals/corr/TzengDHFPLSD15}, the authors apply domain adaptation at the feature level. In \cite{DBLP:journals/corr/TobinFRSZA17} and \cite{conf/iros/MordatchLT15} , the authors used domain and dynamics randomization, respectively. In \cite{DBLP:conf/icml/HigginsPRMBPBBL17}, the authors propose a new multi-stage RL agent, DARLA (DisentAngled Representation Learning Agent), which learns to see before learning to act. More recently, domain adaptation has been studied for robotic manipulators \cite{DBLP:journals/corr/abs-1709-07857,Kalashnikov:2018,james2017transferring,gualtieri2018learning,rusu2017sim,golemo2018sim} in which the authors use raw (pixel) images as state for deep reinforcement learning.

To achieve zero-shot RL requires addressing the uncertainty, un-modeled dynamics, and perception challenges across all three components, namely, agent, environment, and interpreter. There are currently two schools of thought, one focusing on improving dynamics and the other on perception. We argue that the key to achieving robust zero-shot reinforcement learning requires jointly addressing uncertainty in dynamics and variability in perception. 

% There are three main components of reinforcement learning (RL) : 1) Agent, 2) environment, 3) reward generator or interpreter. The agent observes the environment and decides an action to maximize cumulative future rewards. The rewards are generated to drive the agent towards a certain desired behavior. Any variability in agent, environment, or reward generation results in poor transfer of learned policies from simulation-to-simulation (sim2sim) or simulation-to-real world (sim2real). For example, policies learned with a specific plant and actuator configuration (e.g., friction constants) transfer poorly when there is variation in the plant or actuator despite no change in the perception\sravanb{citations}. Policies learned with uncertainty boundaries on plant parameters transfer poorly from one environment to the next when there is variability in perception. 

\begin{figure}[!hbt]
\centering
\includegraphics[width=5 in]{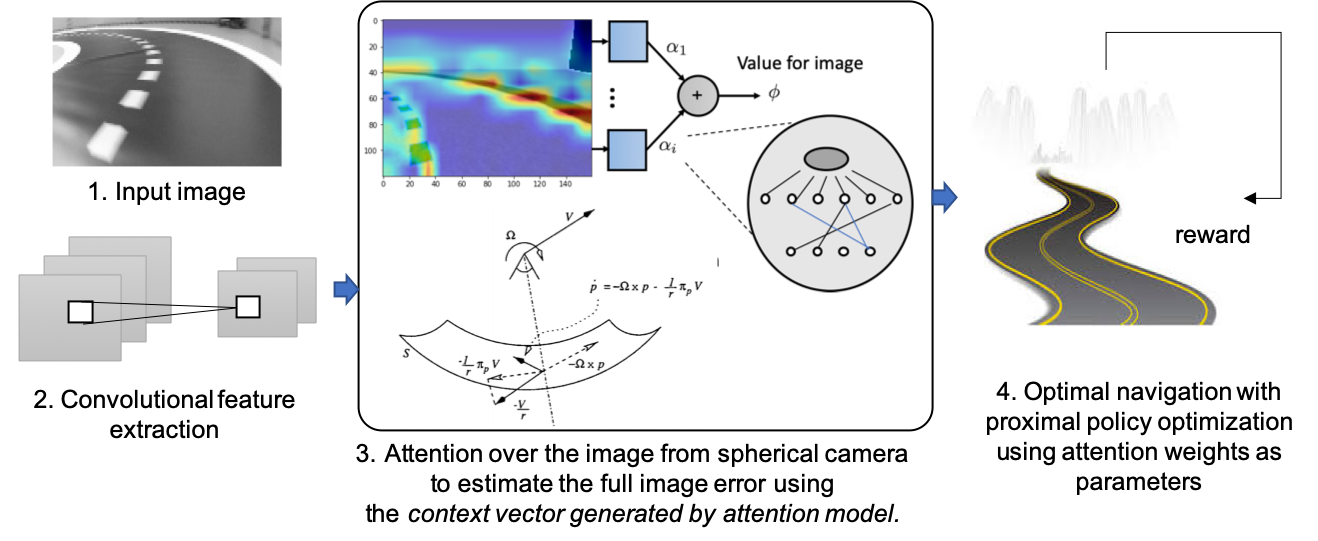}
\caption{The block diagram of our proposed approach using an attention mechanism that preserves passivity-like features of the dynamic system for optimal motion planning. A complete characterization of the class of systems which can be rendered passive from to is beyond the scope of this paper. However, this class is broad, and encompasses mechanical systems modeled by Euler-Lagrange equations.} \label{fig:DACNN}
\end{figure}

We propose a new deep neural network architecture named Deep Attention Convolutional Neural Network (DACNN). An overview of the steps of our proposed approach is shown in Figure~\ref{fig:DACNN}. Our key contribution is that our attention model uniquely captures underlying components in the modern control theoretic approach, i.e.,  image-based servo-ing, without the need for separation of perception and control. The image-based servoing have been succesully applied to robotic control use cases including but not limited to drones. The recent image-based servo-ing methods use image feature vectors which are specific transformation of the raw pixels. We prove that our attention model uniquely captures the image feature vectors, i.e.,  in image-based visual servo control, via annotation vectors. Annotation vectors are extracted from a CNN as described in \cite{bahdanau2014neural}. By defining the image features as annotation vectors, the full image error is defined in term of the weights of the annotation vectors. We assume that the annotation vectors have fixed orientation in the inertial frame. This assumption allows the passivity-like features of the dynamic system to transfer to the full image error when a spherical camera geometry is used. Therefore, we jointly solve the perception and control problem via the attention model that results in robust domain adaptation with zero-shot RL. A complete characterization of the class of systems which can be rendered passive from is beyond the scope of this paper. However, this class is broad, and encompasses mechanical systems modeled by Euler-Lagrange equations \cite{Arcak}.

% Prior methods on improving dynamics integrate H-$infinity$ control with reinforcement learning \cite{Morimoto}. These earlier methods assume that a reasonable model of the plant or actuator can be obtained through well-established system identification techniques \cite{Sim-to-real:Coumans}. Recent methods heavily rely on randomization of plant and actuator parameters in the simulation at the cost of increased learning period \cite{Christensen,Chebotar2018ClosingTS,Gu2016DeepRL}.  

% We organize the paper as follows. First, we provide motivation with comprehensive section on prior and related work. Next, we consider the mathematical formulation of the perception domain transfer problem and describe our approach in the context of control theory and reinforcement learning. To understand the boundaries of our approach to address various perception mismatch modalities in simulation-to-real world, we setup an extensive design of experiments with different textures, track colors, light sources with varying colors, and reward functions. To reduce the entanglement on mismatch in dynamics, we keep the racetrack layout the same across training in simulation and evaluation in real world. 

\section{Our Approach: Attention Models in Optimal Visual Control}\label{sec:approach}

In this section, we describe each constructing block of the overall architecture of our proposed approach in Figure~\ref{fig:DACNN}. First, we describe the core of our architecture, attention neural networks. Second, we describe the underlying assumptions that enable attention networks to perform better than the current state-of-the art approaches. Finally, we provide how autonomous driving physics satisfy the assumptions for deep reinforcement learning with attention networks. 

Our hypothesis is that under certain assumptions on the control system and sensor configuration, a specific type of neural network, i.e., attention network, enables joint perception and stable control of the system even there are significant changes in the environment such as texture and lighting that transforms the observation space. There are several formulation of attention networks for image-captioning and natural language processing. To the best of our knowledge, the attention mechanism in these applications are used in conjunction with recurrent neural networks. There are several formulations of attention models in recurrent neural networks \cite{xu2015show, bahdanau2014neural, Ba2015MultipleOR, DBLP:journals/corr/MnihHGK14}. Attention enables neural networks to "focus" selectively on different parts of the input while generating the corresponding parts of the output sequence. This selective "focus" for a corresponding input is then learned through back-propogation. 

Our formulation is inspired from the attention model in \cite{xu2015show} where the attention-based model can attend to salient part of an image while generating its caption. Intuitively, attention enables the model to see what parts of the image to focus on as it generates a caption. It is very much equivalent to how humans perceive when performing image caption generation or long length machine translation. In the context of autonomous driving, the vehicle needs to focus on the ques from the road such as white and yellow lines but not on the entire road, i.e., grey asphalt, unless there is an obstacle or another vehicle. 

Our main goal is to learn the shortest path on an arbitrarily-drawn track using a ground robot or vehicle as the highest possible speed without going off-the-track, hitting obstacles or other vehicles. We assume that the ground vehicle control is based only on raw (pixel) image and there is no other sensor for position or velocity. In the following, first, we consider the image-based control formulation using construct from the image-based visual servo (IBVS) theory and the kinematic model of our vehicle as defined in \cite{Borelli:2015}. We provide the necessary conditions required to preserve passivity-like properties. Second, we consider the full image error for the control problem. We propose that the image features and full image error is defined in terms of \textit{annotation vectors} and their corresponding weights. We show that the proposed formulation guarantees a stabilizing controller. Finally, we describe the learning task based on attention model. 

The image-based only IBVS approach for our ground vehicle problem can be solved if and only if the image geometry is spherical. In the classical setting, the state of our vehicle is defined as $s = [x, y, \psi, v]$ where $(x, y)$ is the position of the vehicle on a 2D plane, $\psi$ is the steering angle of the center of the mass, and $v$ is the velocity of the vehicle in 2D plane. The state transition function is defined by the discrete-time formulation of the kinematic model in Equations~\ref{eq:x}-\ref{eq:a} as formulated in \cite{Borelli:2015}:
\begin{eqnarray}
\dot{x} & = & v \cos(\psi + \beta) \label{eq:x}\\
\dot{y} & = & v \sin(\psi + \beta)\\
\dot{\psi} & = & (v/l_r) \sin(\beta)\\
\dot{v} & = & a \label{eq:a}\\
\beta & = & \tan^{-1} \left(({l_r}/{L}) \tan (\delta_f) \right)
\end{eqnarray}
$l_r$ is the distance of the center of mass to the rear axle, $L$ is the length of the vehicle, $a$ is the acceleration, and $\delta_f$ is the steering of the front vehicle\footnote{The rear wheels are fixed and do not steer}. The proposed kinematic performs at par with the dynamic model for model predictive control off-road test using actual-size passenger automobiles. This kinematic model satisfies the passivity property for the control system.

In the classical IBVS formulation, the control state is inferred from the sensor-based observations. That is, the sensor state is the raw-pixel image and a transformation matrix is used to map observations into the control state. In our deep reinforcement learning formulation, we consider observations as the state and the control state transformation is implicitly embedded into the neural network and will not be inferred directly. The deep reinforcement learning algorithm will infer on the control state through a reward function by making trial-and-error based decision on the observation space, i.e., raw pixel images.

In our and IBVS formulation, the observation state is defined as $\mathcal{S}\in \mathbb{R}^L$ where $S = [\mathbf{a}_1, \ldots \mathbf{a}_L] \in \mathcal{S} $ is a matrix and each column $\mathbf{a}_i \in \mathbf{R}^D$ corresponds to a $D$-dimensional feature extracted from the observed image, $L$ and $D$ are finite integers. The geometry of the camera is modeled by its image surface $S$ relative to its focal point. Therefore, the image feature $\mathbf{a_i}$ can be written as a function of is projection $\mathbf{P}_i$ onto the image surface $S$ in the body fixed frame. The image feature in our formulation is the output of the convolutional neural network layers, and the input to the convolutional network layers is the raw-pixel image.
% \begin{equation}
%     \mathbf{a_i} = \mathbf{P}_i / r (\mathbf{P}_i)
% \end{equation}
% where $r$ is the relative depth of the target and the division is element-wise. For a typical camera with a flat image, the relative depth is inversely proportional to the focal length. 

\begin{theorem} \label{thm:spherical}
The passivity-like properties of the body fixed frame dynamics of a rigid object in the image space are preserved if and only if the image geometry is of a spherical camera.
\end{theorem}
The proof of Theorem~\ref{thm:spherical} is in \cite{Hamel:2002}. Our kinematic equations, Equations~\ref{eq:x}-\ref{eq:a}, already exhibits a simple linear cascade system, i.e., $\dot{x} = v R$ and $\dot{v} = a$ where $R$ is a rotation matrix. Since cascade systems exhibit passivity-like properties, to guarantee that the controlled system inherits passivity-like properties, we need to show that the gradient of the full image error contains a skew-symmetric matrix on angular velocities. In the next section, we define the full image error in terms of an attention network.

\begin{figure*}[!hbt]
\centering
\includegraphics[width=5in]{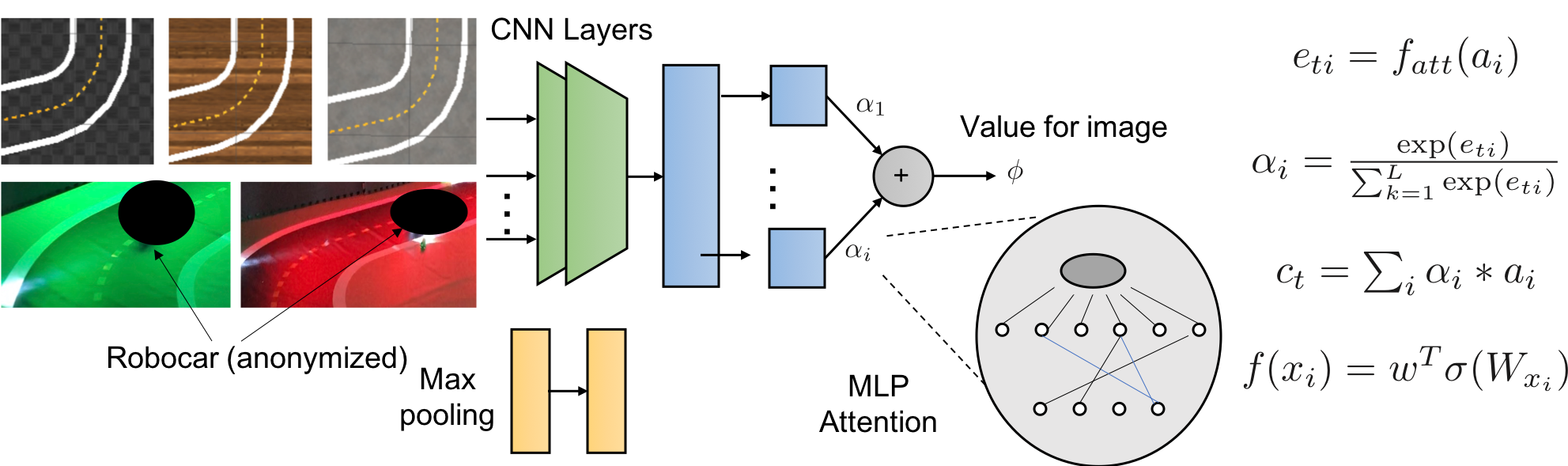}
\caption{The block diagram of the our attention-CNN network.} 
\label{fig:attentionagent}
% \vspace{-10pt}
\end{figure*}

The overall neural network architecture with the deep attention network for our approach is shown in Figure~\ref{fig:attentionagent}. Intuitively, the objective of a visual servo algorithm in image space is to match the observed image to a known ``model'' image of the target. The target is an image of the environment with the desired outcome. In the context of autonomous vehicle, a target is an image of the road where the vehicle is within the white lines and away from obstacles and other vehicles. Our approach does not require a known model image of the target for controls. However, we need to engineer a reward function that will indirectly provide the means to discriminate between desired versus undesired behavior. Therefore, it is necessary to examine the error in the image space. Furthermore, we hypothesize that our attention network model reduces the image space error compared to naively feeding image features extracted from CNN layers. The full image error between the observed and known model image is defined using a combination matrix approach
\begin{equation}
    \delta_1 := C (\mathbf{a}_i - \mathbf{a}^*_i) \label{eq:delta1}
\end{equation}
where $\mathbf{a}^*_i$ are the desired image features and $C$ is the combination matrix that preserve the passivity-like properties. Assuming that $C$ is full rank, we can rewrite the full image error as a weighted sum 
$\delta_1 = \Sigma_{i} \alpha_i (\mathbf{a}_i - \mathbf{a}^*_i)$ where $\alpha_i \geq 0$. The choice of $\alpha_i$ becomes the design component for the control algorithm. We propose that in the above formulation $\mathbf{a}_i$ can be chosen as the annotation vectors in a our modified attention model and $\alpha_i$ are the corresponding weights of the annotation vectors. Suppose that a set of image features, \textit{annotation vectors} are extracted from a CNN as in \cite{xu2015show}. The output of the attention layer from the annotation vectors corresponds to the features extracted at different image locations.  The extractor produces a finite number $L$ of annotation vectors $a = \{\mathbf{a}_1, ..., \mathbf{a}_L\}, \mathbf{a}_i \in \mathbb{R}^D$ where $D$ is a finite integer. These annotation vectors form the state space of our MDP. We define a context vector $\hat{z}$ where $\phi$ is a weighted sum,
\begin{equation}
    \hat{z} = \sum_{i} \alpha_i \mathbf{a}_i 
\end{equation}
For each location $i$, the mechanism generates a positive weight
$\alpha_{i}$ which can be interpreted either as the probability that location $i$ is the right place to focus for producing the control action (the "hard" but stochastic attention mechanism), or as the relative importance to give to location $i$ in weighted sum of the the $a_i$ vectors. The weight $\alpha_{i}$ of each annotation vector $a_i$ is computed by an attention model $f_{att}$. The weight $\alpha_{i}$ is calculated by the softmax function 
%in Equation~\ref{eq:alphati}. 
\begin{equation}
    \alpha_{ti}   =  \exp(e_{ti}) / \Sigma_{k}^L \exp(e_{tk}) \label{eq:alphati}
\end{equation}
where
\begin{equation}
    e_{ti}  =  f_{att} (\mathbf{a}_i, \mathbf{h}_{t-1}), \label{eq:eti}
\end{equation}
$\mathbf{h}_{t-1}$ are hidden state vectors from the previous LSTM cell and $f_{att}$ an attention model. In the literature, there are multiple formulations of the $f_{att}$ model, i.e., additive and multiplicative. In the following, we describe our formulation of  $f_{att}$ based on additive models

The only input to the attention model $f_{att}$ is the annotation vectors. Additive attention (or multi-layer perceptron(MLP) attention) and multiplicative attention (or dot-product attention) are the  most commonly used attention mechanisms. They share the same and unified form of attention introduced above, but are different in how they compute the function $f_{att}$.  We use a modified MLP attention in our network to selectively pick the $a_{i}$ vector for computing the $e_{t_{i}}$, and not have any contextual vector for computing the attention weights. Since the annotation weights $\alpha_i$ are output of a softmax function, $\alpha_i$ are positive. Therefore, we preserve the combination matrix $C$ in Equation~\ref{eq:delta1} to be full rank. This property also ensures that we can stabilize the system if the velocity is available as a control input, i.e., kinematic control. This condition is satisfied by our design. 

While the visual features used as state provide means for robustness to camera and target calibration errors, the rigid-body dynamics of the camera ego-motion are highly coupled when expressed as target motion in the image plane. Therefore, a direct adaptive control approach provide better results. We consider a general formulation of the ground robot navigation as Markov decision process (MDP) and use a vanilla clipped proximate policy optimization algorithm\cite{schulman2017proximal}. In standard policy optimization, the policy to be learned is parametrized by the weights and bias parameters of the underlying neural network. In our case, the underlying neural network contains an attention mechanism in addition to more commonly used convolutional and dense layers. Therefore, the policy optimization solves for the full image error in the visual space while learning the optimal actions for navigation.

% We consider image as a state. This, in general, should lead to a partially-observed MDP. However, our use of the spherical camera geometry and attention model to create a context vector in the form of a visual error function for controller design guarantees passivity-like properties to be preserved as long as the control actions utilize velocity. Therefore, we design our action space to create a velocity controller along with steering instead of position. One caveat is that our ground vehicle is built by off-the-shelf mechanical components with high variances in manufacturing. To alleviate any sim2real issues arising from these variances, we include noise in velocity and steering as described in \cite{BARTO199581} for a general class of navigation problems in reinforcement learning.

\section{Conclusion and Future Work}
Most traditional approaches to control robotic agents rely on extracting features from image or using non-image based sensors for state measurements. The optimal control algorithms that utilize pixel as a state typically need a simulation environment. The control policies trained in a specific simulation environment perform poorly in other environments with the same hardware model configuration. We have tackled two critical problems in our work on domain adaptation. First, we provided a theoretical foundation on the conditions under which joint perception and control will perform as well as the current state-of-the-art or better at lower computational complexity. Second, we implemented our approach for a mobile robot and empirically demonstrated the improvement.

We have proposed a new architecture (DACNN) that strives to attain joint perception and control by learning to focus through the lens of the CNN layers, and achieve zero-shot reinforcement learning, converging to an optimal policy that's transferable across perceptive differences in the environment. The attention model learns to capture the part of the image relevant to driving while the spherical geometry under which the image captures the real-time observation guarantees stable control under the passivity assumption.

We have demonstrated that additive attention can capture the focus required for optimal control theoretically in Section~\ref{sec:approach} and empirically in the context of autonomous-driving in Section~\ref{sec:supp}. This is achieved by designing the context vector in the form of the full error in visual space with respect to the desired visual features. We empirically prove over comprehensive and targeted experiments in simulation and real world that this new mechanism provides a robust domain transfer performance across different textures, colors, and lighting. We have shown that our attention network performs at par or better compared to the current state-of-the-art method based on variational auto-encoders at lower computational complexity and the need to design extensive set of experiments with domain variation. Future work should also look to explore other attention mechanisms like self-attention \cite{vaswani2017attention}, deep siamese attention \cite{wu2018and} to have stronger capabilities to teach focus to the encoded features of our network.
\label{sec:conclusion}

\section{Supplementary Material}\label{sec:supp}
\subsection{Experiments and Results}
In this section, we summarize our experimental results on sim2sim and sim2real. Our main conclusion is two-folds: 1) DACNN provides equivalent or better performance on domain transfer tasks with texture and light variation in the environment and 2) DACNN converges sooner in training compared to deeper neural networks with better performance, i.e., higher total reward, resulting in lower compute cost without degradation. The faster training is achieved by 1) not randomizing the model and environment conditions and 2) allowing the network to focus on jointly optimizing for perception and dynamics. There is no pre-training stage for learning the latent space representation for perception. An in-depth discussion on our experiments and description of baselines adopted are included in the rest of the paper.

\subsubsection{Design of Experiments}
We conducted two sets of experiments with increasing complexity:
\begin{enumerate}[label=Task \Roman* -]
\item \textit{Time trial:} For this task, the robocar is required to complete a given racing track as fast as possible without going off the track. The off-track condition is defined as when all the wheels of the car outside of the race track.
\item \textit{Multi-car racing:} For this task, the robocar is required to complete a given racing track as possible without crashing into two or more bot cars. The crash condition is defined as when the robocar is in the close vicnity of a bot car. The bot car is controlled by the simulation with no learning capability and models the worst case scenario by frequently changing the lanes to block the learner robocar. The number of the bot cars on the track is based on the length of the track.
\end{enumerate}

We consider two sets of domain adaptation tasks in sim2sim and sim2real experiments. First, we use unseen color, texture, and lighting conditions in the evaluation phase and real track environment respectively but we keep the track shape the same in all experiments. Second, we modify the track shape as well as the color and texture.

We consider two baselines in both sim2sim and sim2real transfer experiments. First baseline has vanilla CNN layers to extract features without our attention mechanism. Second baseline is based on the DARLA approach where we use DARLA's $\beta$-VAE neural network to learn a latent state representation and then use the same output layers from our approach to learn the control task. Our implementation of DARLA is as described in \cite{higgins2017darla}, and applied to the autonomous vehicle use case. %We describe our implementation of the second baseline in Section~\ref{sec:DARLA}. 

We trained two baseline models with no domain adaptation, two models with DARLA using different $\beta$ values, two models with DACNN using different number of units, a total of 6 models. We used a vanilla policy optimization implementation and categorical exploration from a widely popular tool kit. We created an interface to our simulation environment which is open-source\footnote{https://github.com/aws-samples/aws-deepracer-workshops/tree/master/Advanced\%20workshops/}. 

\begin{figure*}[!hbt]
\centering
\includegraphics[width=4in]{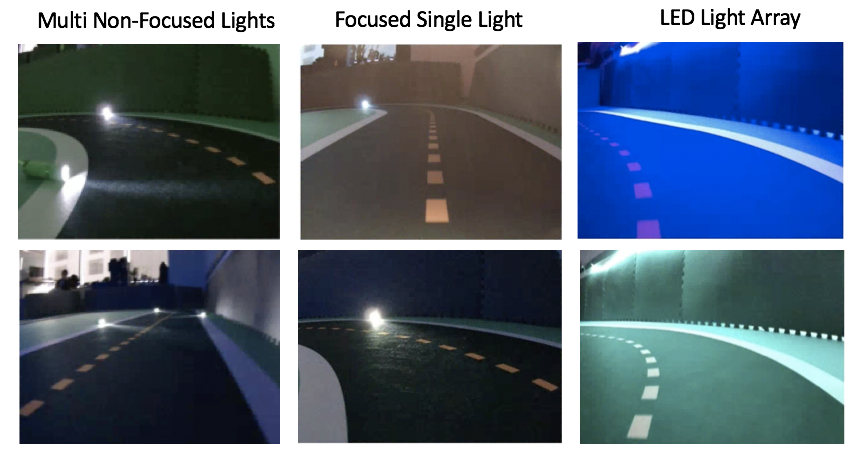}
\caption{The camera capture from robocar's perspective for testing varying sim2real environments.} 
\label{fig:sim2realtest}
% \vspace{-10pt}
\end{figure*}

We performed evaluations on sim2sim and sim2real with textural variants - carpet, wood, concrete, and random lighting effects as seen in Figure~\ref{fig:sim2realtest} with five or more replications. Then, we changed the reward function from penalizing deviations from following center yellow, dotted line to penalizing crossing white lines on the inner and outer edge of the track, and repeated the experiments. 

\subsubsection{Results: Task I - Time-trial}
In total, we have more than 100+ and 20+ tests for sim2sim and sim2real respectively, to reach a statistically significant conclusion. Currently, we have several hundred hours of autonomous driving image, time-series, and event logs from the point-of-view of the car\footnote{See racing competition links at https://driving-olympics.ai/}. In the simulation, we note that the baseline model was never able to finish on the concrete track. However the attention based model finished successfully on all surfaces and all of its converged iterations had significantly higher (80\%+) completion rates. One interesting observation is that our deeper DACNN architecture achieved the highest cumulative reward. We anticipate that the deeper network implicitly extracts the non-linearities from vision-based controls implicitly.

We summarize our sim2real experiment results in Table~\ref{tab:results}. In the real world, we observed DARLA and DACNN performed better than the baselines under lighting variations\footnote{A video of our findings from these experiments are attached to our submission.}. Baseline I uses the probabilistic action space described for ``racetrack problem'' in \cite{BARTO199581} to account for uncertainty in dynamics, while Baseline II uses deterministic action decisions.  We consider two types of DARLA models, one uses data augmentation and the other does not, when training the learn to see model. The data augmentation is performed by various perturbations such as shifting the camera image or adding Gaussian noise. DACNN models use a single layer and two-layer attention layers after vanilla input CNN layers.

\begin{table*}[h]
\caption{Summary of results for simulation-to-simulation and simulation-to-real experiments.}
\centering
%\begin{tabular}{@{}p{0.20\textwidth}*{6}{L{\dimexpr 0.12\textwidth-2\tabcolsep\relax}}@{}}
%{|p{0.20\textwidth}|p{0.20\textwidth}|p{0.20\textwidth}|p{0.20\textwidth}|p{0.20\textwidth}|p{0.20\textwidth}|}
\begin{tabular}{p{0.20\textwidth}p{0.10\textwidth}p{0.10\textwidth}p{0.10\textwidth}p{0.10\textwidth}p{0.10\textwidth}p{0.10\textwidth}}
\toprule & \multicolumn{4}{c}{Simulation-to-Simulation} &\multicolumn{2}{c}{Simulation-to-Reality} \\
\cmidrule(r{4pt}){2-5} \cmidrule(l){6-7}
& Number of iterations to converge & Mean reward at convergence& Success rate for textures & Success rate for lighting &  Success rate for focused light & Success rate for multi non-focused\\
\midrule
Baseline I (uncertainty) \cite{BARTO199581} & 14 & 350 & Fails all & Fails & 0 & 0\\
Baseline II (no adaptation) & 24 & 800 & Fails on concrete & Success & 40\% & 33\%\\
DARLA (augmentation) & 9 & 700 & Success all & Success & N/A & N/A \\
DARLA (no augmentation) & 9 & 600 & Success all & Success & 33\% & 66\%\\
DACNN (shallow) & 12 & 900 & Success all & Success & 66\% & 50\% \\
DACNN (deep) & 11 & 2200 & Success all & Success & in-progress & in-progress \\
\bottomrule
\end{tabular}
\label{tab:results}
\end{table*}

\subsubsection{Results: Task II - Multi-car Racing}
While our results in this task are preliminary, we observed that the reduced computational complexity transferred to the new task of avoiding moving objects. We used a different length and width track to accommodate up to three robocars without crashing into each other. For the same simulation configuration using five and three bot cars changing lanes at specific intervals, the DACNN model converged faster than the deep neural network but slower than the shallow network as shown in Figure~\ref{fig:MultiCarAttention5cars} and Figure~\ref{fig:MultiCarAttention3cars}, respectively. However, the total reward achieved for the DACNN was higher than the shallow one and only slightly higher than the deeper network. The evaluation results on each model during training provided additional insight on the robustness of the DACNN model compared to the deeper network.

\begin{figure}
\centering
\begin{minipage}{.48\textwidth}
  \centering
  \includegraphics[width=2.5in]{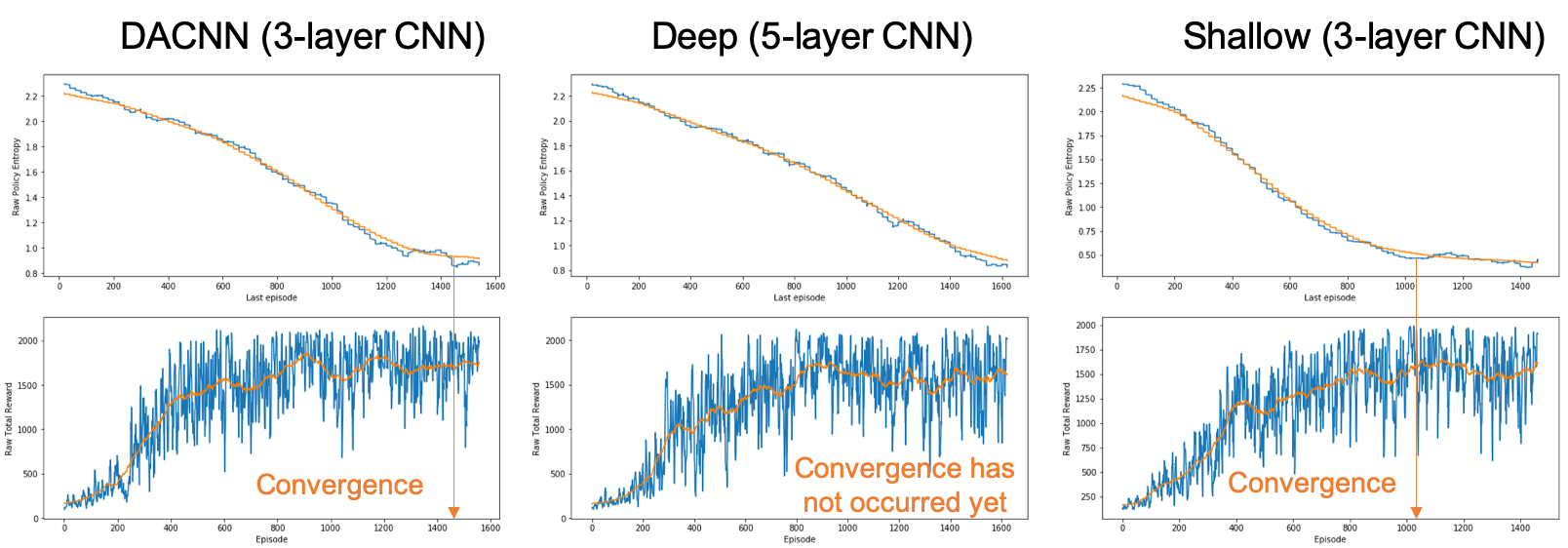}
  \caption{Training entropy and total reward using 5 bot cars where (left) DACNN model with three CNN layers, (center) deep neural network model with five CNN layers but without attention, and (right) shallow neural network with three CNN layer.}
  \label{fig:MultiCarAttention5cars}
\end{minipage}%
\begin{minipage}{.48\textwidth}
  \centering
  \includegraphics[width=2.5in]{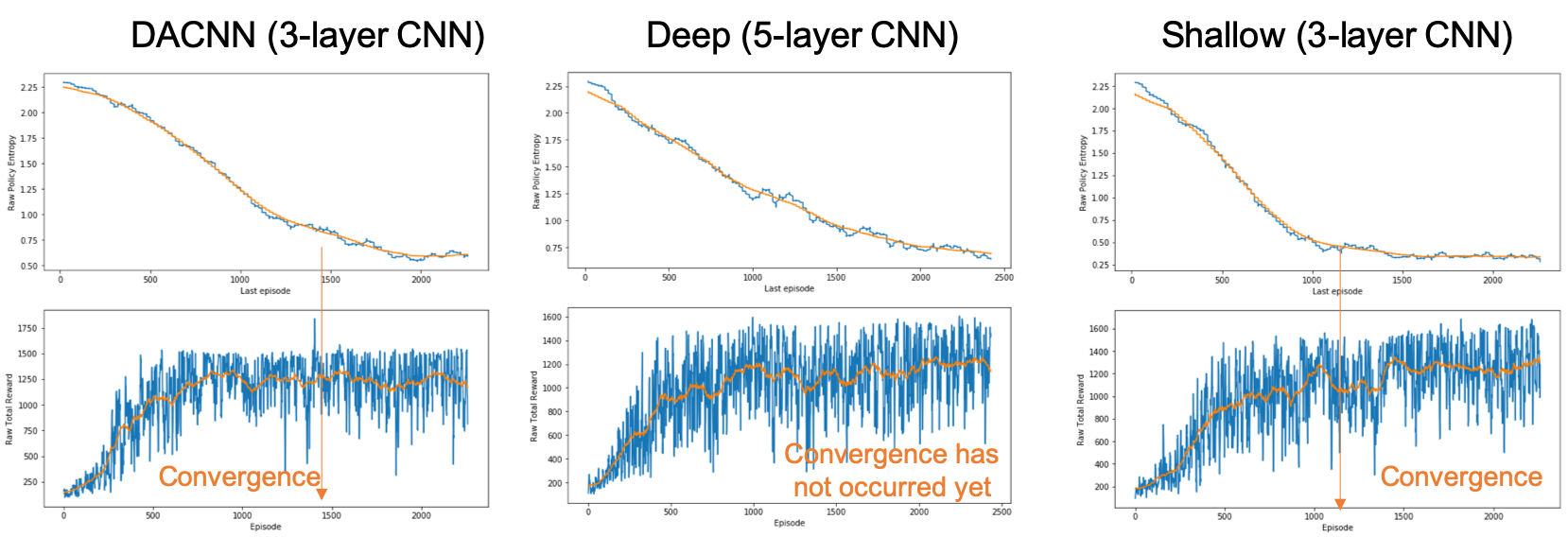}
  \caption{Training entropy and total reward using 3 bot cars where (left) DACNN model with three CNN layers, (center) deep neural network model with five CNN layers but without attention, and (right) shallow neural network with three CNN layer}
  \label{fig:MultiCarAttention5cars}
\end{minipage}
\end{figure}

% \begin{figure}[!hbt]\centering
% \includegraphics[width=3in]{figures/MultiCarAttention5cars.png}
% \caption{Training entropy and total reward using 5 bot cars where (left) DACNN model with three CNN layers, (center) deep neural network model with five CNN layers but without attention, and (right) shallow neural network with three CNN layer.}
% \label{fig:MultiCarAttention5cars}
% \end{figure}

% \begin{figure}[!hbt]\centering
% \includegraphics[width=3in]{figures/MultiCarAttention3cars.png}
% \caption{Training entropy and total reward using 3 bot cars where (left) DACNN model with three CNN layers, (center) deep neural network model with five CNN layers but without attention, and (right) shallow neural network with three CNN layer.}
% \label{fig:MultiCarAttention5cars}
% \end{figure}

\begin{figure}[!hbt]\centering
\includegraphics[width=4in]{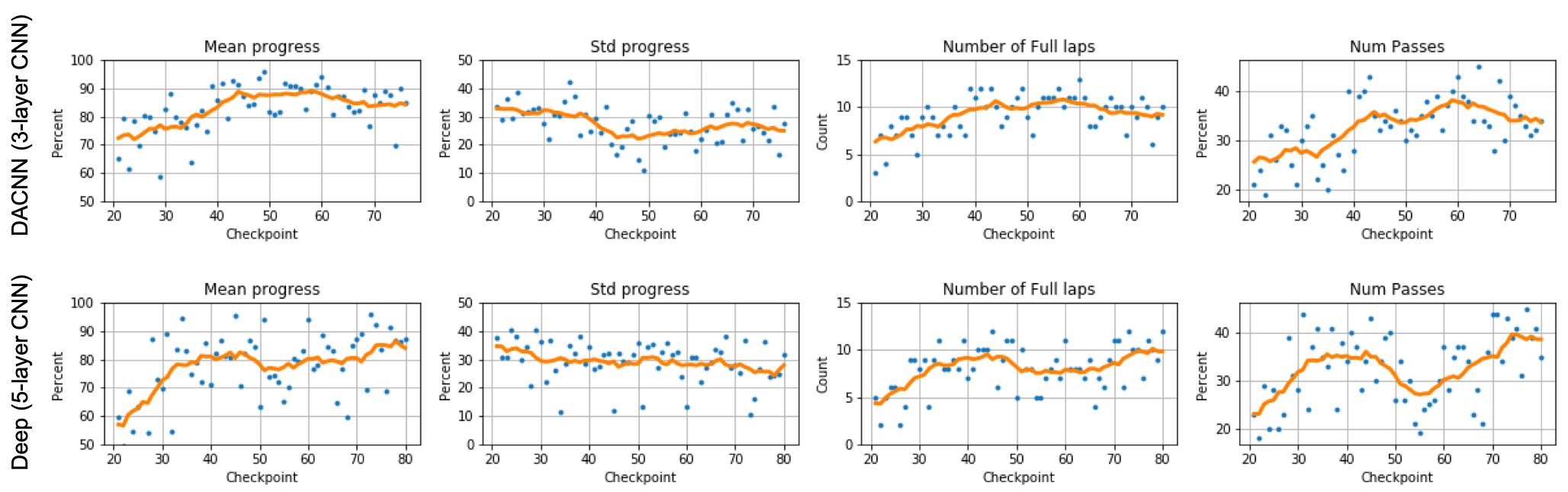}
\caption{The evaluation of models with five bot cars on the track same as the training where (top) DACNN model with three CNN layers plots of evaluation and (bottom) deep neural network with three CNN layer but without attention model plots of evaluation.}
\label{fig:multicar_attention_evaluation}
\end{figure}

Next, we consider evaluation of the five-bot car model on the same track as the training and of the three-bot model on a more complicated track shape with varying textures. The evaluation results for the five-bot car model are shown in Figure~\ref{fig:multicar_attention_evaluation}. We provide the statistics for mean progress over 15 evaluations across each checkpoint model from the training job in the first column of Figure~\ref{fig:multicar_attention_evaluation}. The DACNN model starts achieving more than 70\% progress around the track early on. The variation across evaluations of the same model is shown in the second column. The DACNN model has tighter variation plots. Similarly for the number of cars passed during evaluation shown in the fourth column, the DACNN model consistently performs more than 30 passes after iteration 40. The evaluation of the three-bot car model on a track with a more complicated track shape and several texture changes including concrete, carpet, and wood revealed the limits of both the DACNN and deeper network models. None of the models were able to finish the track yet the DACNN model was able to complete more full laps than the deeper model with higher number of passes.

% \begin{figure}[!hbt]\centering
% \includegraphics[width=5in]{figures/multicar_attention_evaluation_3car.png}
% \caption{The evaluation of models with three bot cars on a track with varying shape and texture than the training track where (top) DACNN model with three CNN layers plots of evaluation and (bottom) deep neural network with three CNN layer but without attention model plots of evaluation.}
% \label{fig:multicar_attention_evaluation_3car}
% \end{figure}

\subsubsection{Discussion}

We use Gradient-weighted Class Activation Mapping (Grad-CAM) in \cite{selvaraju2017grad} to visualize the impact of our baseline versus proposed approach on the image space prior to the output layers for control. Grad-CAM applies to CNNs used in reinforcement learning, without any architectural changes or re-training. Grad-CAM uses the gradients of any target concept, flowing into the final convolutional layer to produce a coarse localization map highlighting the important regions in the image for predicting the concept.

\begin{figure}[!hbt]\centering
\includegraphics[width=3in]{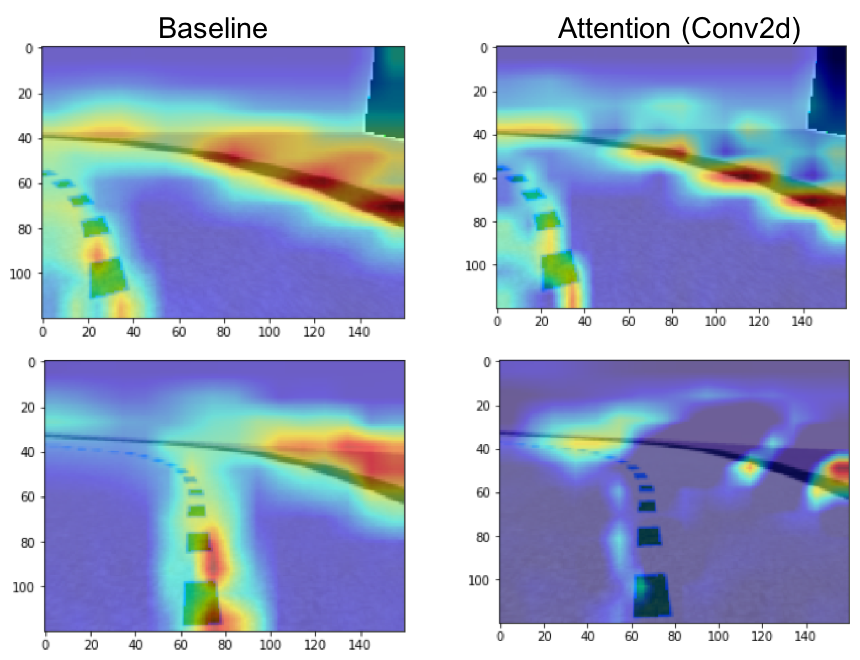}
\caption{(Left) Baseline model on the real-world data from the robocar's perspective is distracted by objects and surfaces outside of the racing track. (Right) DACNN focuses on the current and next actions without distractions to stay near the center of the track.}
\label{fig:gradcam_dacnn}
\end{figure}

In Figure~\ref{fig:gradcam_dacnn}, we compared our baseline and basic DACNN model on the same image collected from the real world through the robocar's perspective. The warmer colors (yellow-red) correspond to focus areas where cooler colors (blue-purple) correspond to ignored areas. It is clear that the DACNN learns to focus on the track and not distracted by objects and surfaces outside the track. Note that both models are saved at the same iteration so that we can observe whether attention outperforms the baseline at same computational training step. 

\begin{figure}[!hbt]\centering
\includegraphics[width=3.5in]{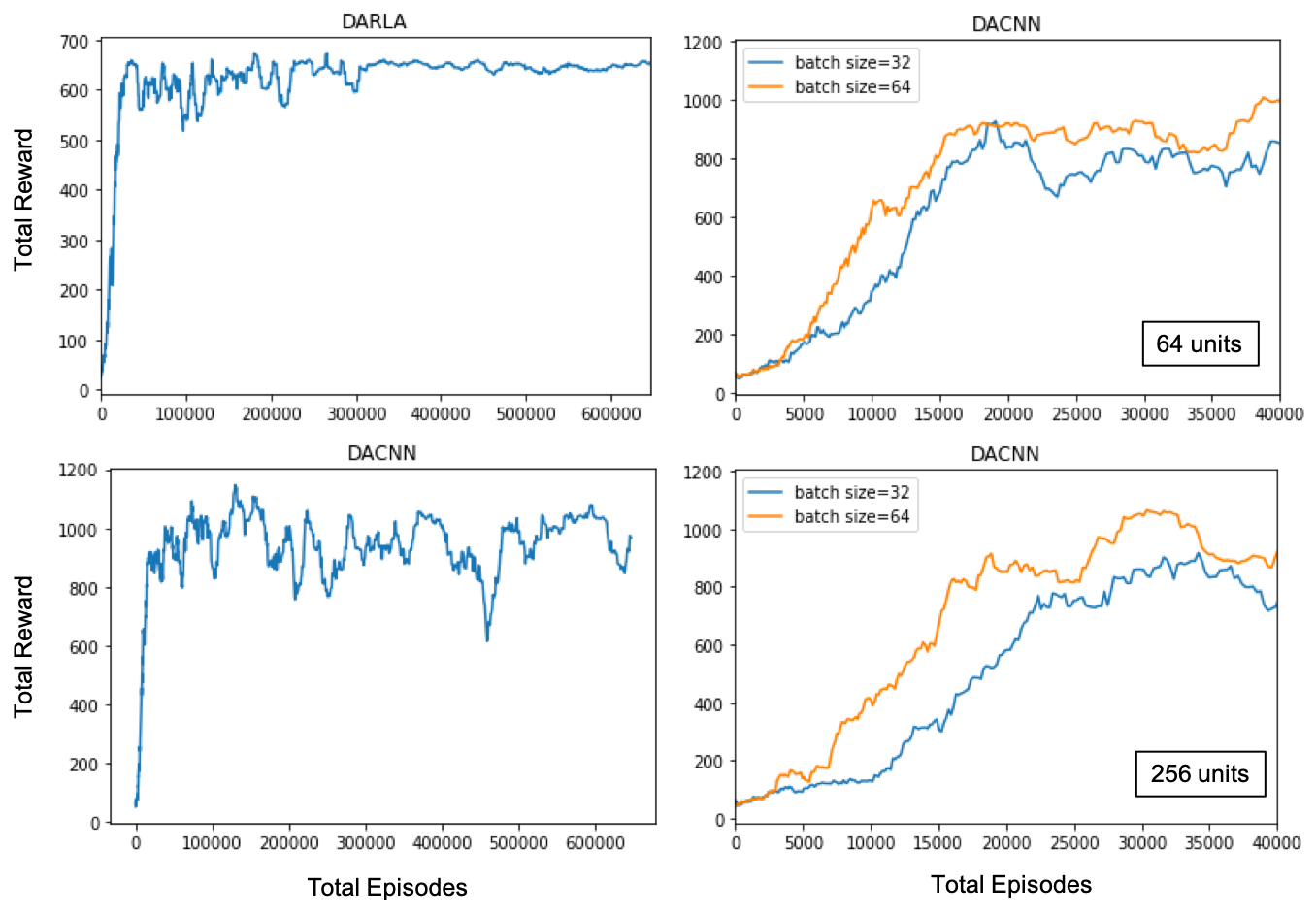}
\caption{(Top Left) The rewards accumulated during the training of the DARLA agent and (Bottom Left) the rewards accumulated during the training of our attention model with the clipped proximal policy optimization \cite{schulman2017proximal} and categorical. The impact of hyper-parameter variation, i.e., batch size and attention units, on learning is shown on the right.}
\label{fig:rewards}
\end{figure}

Figure~\ref{fig:rewards} compare the rewards accumulating for DARLA (top left) and our attention model (bottom left) during training with mini-batch size of 64. The reward function incentives for staying close to the yellow, dotted-line and higher speeds while penalizing for getting out of the track ans steering too frequently. For our basic DACNN models, we observe that the algorithm starts to converge around the same time as DARLA after accounting to DARLA's pre-training period. Therefore, the training performance is maintained. The impact of changing the batch size from 64 to 32 for 64 unit (top right) attention versus a granular 256 unit (bottom right) attention model is shown on the right in Figure~\ref{fig:rewards}. The increased batch size results in faster learning.

% sahika to add a figure here now from experiments

\subsection{Related Work}\label{sec:background} % \sahika{Added comparisons to our work}
\textbf{Vision-based servo control}.  In this paper, we consider image-based optimal control which is an extension of image-based visual servo (IBVS) control.  Classic IBVS was developed for serial-link robotic manipulators \cite{Sanderson:87,Rives:92} and aims to control the dynamics of features in the image plane directly \cite{Hutchinson:96, Hutchinson:2007}. More recent work on visual servoing focused on unmanned aerial vehicles (see \cite{Lu:2018} and reference therein). Our work differs from visual servoing most commonly used in unmanned aerial vehicles because we do not have a separate motion planning module. Our work is most similar to recent work on robotic manipulators \cite{DBLP:journals/corr/abs-1709-07857,Kalashnikov:2018} in which the authors use raw (pixel) images as state for deep reinforcement learning. IBVS methods offer advantages in robustness to camera and target calibration errors, reduced computational complexity. One caveat of the classical IBVS is that it is necessary to determine the depth of each visual feature used in the image error criterion independently from the control algorithm. One of the approaches to overcome this issue is to use adaptive control, hence, the motivation to use reinforcement learning as a direct adaptive control method. 

\textbf{Domain Adaptation for Robot Learning.} In domain adaptation literature for robot learning, our approach is comparable to \cite{DBLP:conf/icml/HigginsPRMBPBBL17} where the authors propose a new multi-stage RL agent, DARLA (DisentAngled Representation Learning Agent), which learns to see before learning to act. In our approach, we do not separate perception from dynamics but our intuition to create a latent \textit{attention} space for dynamics is a common theme. Our approach differs from recent work on robotic manipulators because the state is based on image only and not augmented with other control state information such as position. Moreover, the use of attention network to create a latent space is new. 
In \cite{DBLP:journals/corr/TzengDHFPLSD15}, the authors apply domain adaptation at the feature level. In \cite{DBLP:journals/corr/TobinFRSZA17} and \cite{conf/iros/MordatchLT15}, the authors use domain and dynamics randomization, respectively. 

\textbf{Attention Mechanisms in Reinforcement Learning.} Attention models were applied with remarkable success to complex visual tasks such as image captioning \cite{xu2015show} and machine translation \cite{bahdanau2014neural}. However, attention models have mostly been applied to recurrent neural networks and not for optimal visual-servoing tasks. In \cite{MnihHGK14} and \cite{Liang2019}, the authors use a recurrent neural network (RNN) which processes inputs sequentially and incrementally combines information to build up a dynamic internal representation of the scene or environment. We hypothesize that convolutional neural network (CNN) layers can capture local dependencies needed to create an approximate model for the optimal control task. Instead, our approach passes images sampled from a scene through multiple CNN layers prior to the attention network, hence, has the additional advantage of invariance to lighting and texture inherent in CNNs \cite{LeCun:1998}. In other previous attempts to integrate attention with RL, the authors have largely used hand-crafted features as inputs to the attention model \cite{Manchin:2019}. The hand-crafted features require a large number of hyper-parameters and are not invariant to lighting and texture. Our CNN-based attention network overcomes these challenges in lighting and texture. While it is possible to \textit{segment} focus areas separately as described in \cite{Choi:2017}, the delay caused by the model inference is too large to construct a stable controller.

\bibliography{MAIN}

\begin{thebibliography}{10}
\providecommand{\url}[1]{#1}
\csname url@rmstyle\endcsname
\providecommand{\newblock}{\relax}
\providecommand{\bibinfo}[2]{#2}
\providecommand\BIBentrySTDinterwordspacing{\spaceskip=0pt\relax}
\providecommand\BIBentryALTinterwordstretchfactor{4}
\providecommand\BIBentryALTinterwordspacing{\spaceskip=\fontdimen2\font plus
\BIBentryALTinterwordstretchfactor\fontdimen3\font minus
  \fontdimen4\font\relax}
\providecommand\BIBforeignlanguage[2]{{%
\expandafter\ifx\csname l@#1\endcsname\relax
\typeout{** WARNING: IEEEtran.bst: No hyphenation pattern has been}%
\typeout{** loaded for the language `#1'. Using the pattern for}%
\typeout{** the default language instead.}%
\else
\language=\csname l@#1\endcsname
\fi
#2}}

\bibitem{Sim-to-real:Coumans}
\BIBentryALTinterwordspacing
J.~Tan, T.~Zhang, E.~Coumans, A.~Iscen, Y.~Bai, D.~Hafner, S.~Bohez, and
  V.~Vanhoucke, ``Sim-to-real: Learning agile locomotion for quadruped
  robots,'' \emph{CoRR}, vol. abs/1804.10332, 2018. [Online]. Available:
  \url{http://arxiv.org/abs/1804.10332}
\BIBentrySTDinterwordspacing

\bibitem{DBLP:journals/corr/TzengDHFPLSD15}
\BIBentryALTinterwordspacing
E.~Tzeng, C.~Devin, J.~Hoffman, C.~Finn, X.~Peng, S.~Levine, K.~Saenko, and
  T.~Darrell, ``Towards adapting deep visuomotor representations from simulated
  to real environments,'' \emph{CoRR}, vol. abs/1511.07111, 2015. [Online].
  Available: \url{http://arxiv.org/abs/1511.07111}
\BIBentrySTDinterwordspacing

\bibitem{DBLP:journals/corr/TobinFRSZA17}
\BIBentryALTinterwordspacing
J.~Tobin, R.~Fong, A.~Ray, J.~Schneider, W.~Zaremba, and P.~Abbeel, ``Domain
  randomization for transferring deep neural networks from simulation to the
  real world,'' \emph{CoRR}, vol. abs/1703.06907, 2017. [Online]. Available:
  \url{http://arxiv.org/abs/1703.06907}
\BIBentrySTDinterwordspacing

\bibitem{conf/iros/MordatchLT15}
I.~Mordatch, K.~Lowrey, and E.~Todorov, ``Ensemble-cio: Full-body dynamic
  motion planning that transfers to physical humanoids.'' in \emph{IROS}.\hskip
  1em plus 0.5em minus 0.4em\relax IEEE, 2015, pp. 5307--5314.

\bibitem{DBLP:conf/icml/HigginsPRMBPBBL17}
\BIBentryALTinterwordspacing
I.~Higgins, A.~Pal, A.~A. Rusu, L.~Matthey, C.~Burgess, A.~Pritzel,
  M.~Botvinick, C.~Blundell, and A.~Lerchner, ``{DARLA:} improving zero-shot
  transfer in reinforcement learning,'' in \emph{Proceedings of the 34th
  International Conference on Machine Learning, {ICML} 2017, Sydney, NSW,
  Australia, 6-11 August 2017}, 2017, pp. 1480--1490. [Online]. Available:
  \url{http://proceedings.mlr.press/v70/higgins17a.html}
\BIBentrySTDinterwordspacing

\bibitem{DBLP:journals/corr/abs-1709-07857}
\BIBentryALTinterwordspacing
K.~Bousmalis, A.~Irpan, P.~Wohlhart, Y.~Bai, M.~Kelcey, M.~Kalakrishnan,
  L.~Downs, J.~Ibarz, P.~Pastor, K.~Konolige, S.~Levine, and V.~Vanhoucke,
  ``Using simulation and domain adaptation to improve efficiency of deep
  robotic grasping,'' \emph{CoRR}, vol. abs/1709.07857, 2017. [Online].
  Available: \url{http://arxiv.org/abs/1709.07857}
\BIBentrySTDinterwordspacing

\bibitem{Kalashnikov:2018}
\BIBentryALTinterwordspacing
D.~Kalashnikov, A.~Irpan, P.~Pastor, J.~Ibarz, A.~Herzog, E.~Jang, D.~Quillen,
  E.~Holly, M.~Kalakrishnan, V.~Vanhoucke, and S.~Levine, ``Qt-opt: Scalable
  deep reinforcement learning for vision-based robotic manipulation,''
  \emph{CoRR}, vol. abs/1806.10293, 2018. [Online]. Available:
  \url{http://arxiv.org/abs/1806.10293}
\BIBentrySTDinterwordspacing

\bibitem{james2017transferring}
S.~James, A.~J. Davison, and E.~Johns, ``Transferring end-to-end visuomotor
  control from simulation to real world for a multi-stage task,'' \emph{arXiv
  preprint arXiv:1707.02267}, 2017.

\bibitem{gualtieri2018learning}
M.~Gualtieri and R.~Platt, ``Learning 6-dof grasping and pick-place using
  attention focus,'' in \emph{Conference on Robot Learning}, 2018, pp.
  477--486.

\bibitem{rusu2017sim}
A.~A. Rusu, M.~Ve{\v{c}}er{\'\i}k, T.~Roth{\"o}rl, N.~Heess, R.~Pascanu, and
  R.~Hadsell, ``Sim-to-real robot learning from pixels with progressive nets,''
  in \emph{Conference on Robot Learning}, 2017, pp. 262--270.

\bibitem{golemo2018sim}
F.~Golemo, A.~A. Taiga, A.~Courville, and P.-Y. Oudeyer, ``Sim-to-real transfer
  with neural-augmented robot simulation,'' in \emph{Conference on Robot
  Learning}, 2018, pp. 817--828.

\bibitem{bahdanau2014neural}
D.~Bahdanau, K.~Cho, and Y.~Bengio, ``Neural machine translation by jointly
  learning to align and translate,'' \emph{arXiv preprint arXiv:1409.0473},
  2014.

\bibitem{Arcak}
M.~{Arcak}, ``Passivity as a design tool for group coordination,'' \emph{IEEE
  Transactions on Automatic Control}, vol.~52, no.~8, pp. 1380--1390, Aug 2007.

\bibitem{xu2015show}
K.~Xu, J.~Ba, R.~Kiros, K.~Cho, A.~Courville, R.~Salakhutdinov, R.~Zemel, and
  Y.~Bengio, ``Show, attend and tell: Neural image caption generation with
  visual attention,'' \emph{arXiv preprint arXiv:1502.03044}, 2015.

\bibitem{Ba2015MultipleOR}
J.~Ba, V.~Mnih, and K.~Kavukcuoglu, ``Multiple object recognition with visual
  attention,'' \emph{CoRR}, vol. abs/1412.7755, 2015.

\bibitem{DBLP:journals/corr/MnihHGK14}
\BIBentryALTinterwordspacing
V.~Mnih, N.~Heess, A.~Graves, and K.~Kavukcuoglu, ``Recurrent models of visual
  attention,'' \emph{CoRR}, vol. abs/1406.6247, 2014. [Online]. Available:
  \url{http://arxiv.org/abs/1406.6247}
\BIBentrySTDinterwordspacing

\bibitem{Borelli:2015}
J.~{Kong}, M.~{Pfeiffer}, G.~{Schildbach}, and F.~{Borrelli}, ``Kinematic and
  dynamic vehicle models for autonomous driving control design,'' in \emph{2015
  IEEE Intelligent Vehicles Symposium (IV)}, June 2015, pp. 1094--1099.

\bibitem{Hamel:2002}
T.~{Hamel} and R.~{Mahony}, ``Visual servoing of an under-actuated dynamic
  rigid-body system: an image-based approach,'' \emph{IEEE Transactions on
  Robotics and Automation}, vol.~18, no.~2, pp. 187--198, April 2002.

\bibitem{schulman2017proximal}
J.~Schulman, F.~Wolski, P.~Dhariwal, A.~Radford, and O.~Klimov, ``Proximal
  policy optimization algorithms,'' \emph{arXiv preprint arXiv:1707.06347},
  2017.

\bibitem{vaswani2017attention}
A.~Vaswani, N.~Shazeer, N.~Parmar, J.~Uszkoreit, L.~Jones, A.~N. Gomez,
  {\L}.~Kaiser, and I.~Polosukhin, ``Attention is all you need,'' in
  \emph{Advances in neural information processing systems}, 2017, pp.
  5998--6008.

\bibitem{wu2018and}
L.~Wu, Y.~Wang, J.~Gao, and X.~Li, ``Where-and-when to look: Deep siamese
  attention networks for video-based person re-identification,'' \emph{IEEE
  Transactions on Multimedia}, 2018.

\bibitem{higgins2017darla}
I.~Higgins, A.~Pal, A.~Rusu, L.~Matthey, C.~Burgess, A.~Pritzel, M.~Botvinick,
  C.~Blundell, and A.~Lerchner, ``Darla: Improving zero-shot transfer in
  reinforcement learning,'' in \emph{Proceedings of the 34th International
  Conference on Machine Learning-Volume 70}.\hskip 1em plus 0.5em minus
  0.4em\relax JMLR. org, 2017, pp. 1480--1490.

\bibitem{BARTO199581}
\BIBentryALTinterwordspacing
A.~G. Barto, S.~J. Bradtke, and S.~P. Singh, ``Learning to act using real-time
  dynamic programming,'' \emph{Artificial Intelligence}, vol.~72, no.~1, pp. 81
  -- 138, 1995. [Online]. Available:
  \url{http://www.sciencedirect.com/science/article/pii/000437029400011O}
\BIBentrySTDinterwordspacing

\bibitem{selvaraju2017grad}
R.~R. Selvaraju, M.~Cogswell, A.~Das, R.~Vedantam, D.~Parikh, and D.~Batra,
  ``Grad-{CAM}: Visual explanations from deep networks via gradient-based
  localization,'' in \emph{Proceedings of the IEEE International Conference on
  Computer Vision}, 2017, pp. 618--626.

\bibitem{Sanderson:87}
L.~{Weiss}, A.~{Sanderson}, and C.~{Neuman}, ``Dynamic sensor-based control of
  robots with visual feedback,'' \emph{IEEE Journal on Robotics and
  Automation}, vol.~3, no.~5, pp. 404--417, October 1987.

\bibitem{Rives:92}
B.~{Espiau}, F.~{Chaumette}, and P.~{Rives}, ``A new approach to visual
  servoing in robotics,'' \emph{IEEE Transactions on Robotics and Automation},
  vol.~8, no.~3, pp. 313--326, June 1992.

\bibitem{Hutchinson:96}
S.~{Hutchinson}, G.~D. {Hager}, and P.~I. {Corke}, ``A tutorial on visual servo
  control,'' \emph{IEEE Transactions on Robotics and Automation}, vol.~12,
  no.~5, pp. 651--670, Oct 1996.

\bibitem{Hutchinson:2007}
F.~{Chaumette} and S.~{Hutchinson}, ``Visual servo control. ii. advanced
  approaches [tutorial],'' \emph{IEEE Robotics Automation Magazine}, vol.~14,
  no.~1, pp. 109--118, March 2007.

\bibitem{Lu:2018}
Y.~Lu, Z.~Xue, G.-S. Xia, and L.~Zhang, ``A survey on vision-based uav
  navigation,'' \emph{Geo-spatial Information Science}, vol.~21, no.~1, pp.
  21--32, 2018.

\bibitem{MnihHGK14}
\BIBentryALTinterwordspacing
V.~Mnih, N.~Heess, A.~Graves, and K.~Kavukcuoglu, ``Recurrent models of visual
  attention,'' \emph{CoRR}, vol. abs/1406.6247, 2014. [Online]. Available:
  \url{http://arxiv.org/abs/1406.6247}
\BIBentrySTDinterwordspacing

\bibitem{Liang2019}
\BIBentryALTinterwordspacing
X.~Liang, Q.~Wang, Y.~Feng, Z.~Liu, and J.~Huang, ``{VMAV-C:} {A} deep
  attention-based reinforcement learning algorithm for model-based control,''
  \emph{CoRR}, vol. abs/1812.09968, 2018. [Online]. Available:
  \url{http://arxiv.org/abs/1812.09968}
\BIBentrySTDinterwordspacing

\bibitem{LeCun:1998}
\BIBentryALTinterwordspacing
Y.~LeCun and Y.~Bengio, ``The handbook of brain theory and neural networks,''
  M.~A. Arbib, Ed.\hskip 1em plus 0.5em minus 0.4em\relax Cambridge, MA, USA:
  MIT Press, 1998, ch. Convolutional Networks for Images, Speech, and Time
  Series, pp. 255--258. [Online]. Available:
  \url{http://dl.acm.org/citation.cfm?id=303568.303704}
\BIBentrySTDinterwordspacing

\bibitem{Manchin:2019}
\BIBentryALTinterwordspacing
A.~Manchin, E.~Abbasnejad, and A.~van~den Hengel, ``Reinforcement learning with
  attention that works: {A} self-supervised approach,'' \emph{CoRR}, vol.
  abs/1904.03367, 2019. [Online]. Available:
  \url{http://arxiv.org/abs/1904.03367}
\BIBentrySTDinterwordspacing

\bibitem{Choi:2017}
\BIBentryALTinterwordspacing
J.~Choi, B.~Lee, and B.~Zhang, ``Multi-focus attention network for efficient
  deep reinforcement learning,'' \emph{CoRR}, vol. abs/1712.04603, 2017.
  [Online]. Available: \url{http://arxiv.org/abs/1712.04603}
\BIBentrySTDinterwordspacing

\end{thebibliography}
\bibliographystyle{IEEEtran}

%
%\subsubsection*{Acknowledgments}
%
%Use unnumbered third level headings for the acknowledgments. All acknowledgments
%go at the end of the paper. Do not include acknowledgments in the anonymized
%submission, only in the final paper.
%
%\section*{References}
%
%References follow the acknowledgments. Use unnumbered first-level heading for
%the references. Any choice of citation style is acceptable as long as you are
%consistent. It is permissible to reduce the font size to \verb+small+ (9 point)
%when listing the references. {\bf Remember that you can use more than eight
%  pages as long as the additional pages contain \emph{only} cited references.}
%\medskip
%
%\small
%
%[1] Alexander, J.A.\ \& Mozer, M.C.\ (1995) Template-based algorithms for
%connectionist rule extraction. In G.\ Tesauro, D.S.\ Touretzky and T.K.\ Leen
%(eds.), {\it Advances in Neural Information Processing Systems 7},
%pp.\ 609--616. Cambridge, MA: MIT Press.
%
%[2] Bower, J.M.\ \& Beeman, D.\ (1995) {\it The Book of GENESIS: Exploring
%  Realistic Neural Models with the GEneral NEural SImulation System.}  New York:
%TELOS/Springer--Verlag.
%
%[3] Hasselmo, M.E., Schnell, E.\ \& Barkai, E.\ (1995) Dynamics of learning and
%recall at excitatory recurrent synapses and cholinergic modulation in rat
%hippocampal region CA3. {\it Journal of Neuroscience} {\bf 15}(7):5249-5262.

\end{document}